\documentclass[10pt,twocolumn,letterpaper]{article}

\usepackage{iccv}
\usepackage{times}
\usepackage{epsfig}
\usepackage{graphicx}
\usepackage{amsmath}
\usepackage{amssymb}

\usepackage[breaklinks=true,bookmarks=false]{hyperref}

\iccvfinalcopy 

\begin{document}

\title{Deepfake Video Detection Using Convolutional Vision Transformer}

\author{Deressa Wodajo\\
Jimma University\\
{\tt\small deressa.wodajo@ju.edu.et}
\and
Solomon Atnafu\\
Addis Ababa University\\
{\tt\small solomon.atnafu@aau.edu.et}
}

\maketitle
\ificcvfinal\thispagestyle{empty}\fi

\begin{abstract}

   The rapid advancement of deep learning models that can 
   generate and synthesis hyper-realistic videos known as 
   Deepfakes and their ease of access have raised concern 
   on possible malicious intent use. Deep learning techniques 
   can now generate faces, swap faces between two subjects 
   in a video, alter facial expressions, change gender, and 
   alter facial features, to list a few. These powerful video 
   manipulation methods have potential use in many fields. 
   However, they also pose a looming threat to everyone if 
   used for harmful purposes such as identity theft, phishing, 
   and scam. In this work, we propose a Convolutional Vision 
   Transformer for the detection of Deepfakes. 
   The Convolutional Vision Transformer has two components: 
   Convolutional Neural Network (CNN) and Vision 
   Transformer (ViT). The CNN extracts learnable features 
   while the ViT takes in the learned features as input and 
   categorizes them using an attention mechanism. 
   We trained our model on the DeepFake Detection 
   Challenge Dataset (DFDC) and have achieved 91.5 
   percent accuracy, an AUC value of 0.91, and a loss value 
   of 0.32. Our contribution is that we have added a CNN 
   module to the ViT architecture and have achieved a 
   competitive result on the DFDC dataset.

\end{abstract}

\section{Introduction}

Technologies for altering images, videos, and audios are 
developing rapidly~\cite{Authors01,Authors02}. 
Techniques and technical expertise to create and 
manipulate digital content are also easily accessible. 
Currently, it is possible to seamlessly generate 
hyper-realistic digital images~\cite{Authors03} with a 
little resource and an easy how-to-do instructions available 
online~\cite{Authors04,Authors05}. Deepfake is a 
technique which aims to replace the face of a targeted 
person by the face of someone else in a 
video~\cite{Authors06}. It is created by splicing
synthesized face region into the original image
~\cite{Authors02}. The term can also mean to represent 
the final output of a hype-realistic video created. Deepfakes 
can be used for creation of hyper-realistic Computer 
Generated Imagery (CGI), Virtual Reality (VR)
~\cite{Authors07}, Augmented Reality (AR), 
Education, Animation, Arts, and Cinema~\cite{Authors08}. 
However, since Deepfakes are deceptive in nature, they can 
also be used for malicious purposes.

Since the Deepfake phenomenon, various authors have 
proposed different mechanisms to differentiate real videos 
from fake ones. As pointed by~\cite{Authors09}, even 
though each proposed mechanism has its strength, current 
detection methods lack generalizability. The authors noted 
that current existing models focus on the Deepfake creation 
tools to tackle by studying their supposed behaviors. 
For instance, Yuezun \etal~\cite{Authors10} and 
TackHyun \etal~\cite{Authors11} used inconsistencies 
in eye blinking to detect Deepfakes. However, using the work 
of Konstantinos \etal~\cite{Authors12} and 
Hai \etal~\cite{Authors13}, it is now possible to mimic 
eye blinking. The authors in~\cite{Authors12} presented a 
system that generates videos of talking heads with natural facial 
expressions such as eye blinking. The authors in~\cite{Authors13} 
proposed a model that can generate facial expression from 
a portrait. Their system can synthesis a still picture to express 
emotions, including a hallucination of eye-blinking motions.

We base our work on two weaknesses of Deepfake detection 
methods pointed out by~\cite{Authors09,Authors14}: 
data preprocessing, and generality. Polychronis 
\etal~\cite{Authors14} noted that current Deepfake
detection systems focus mostly on presenting their proposed 
architecture, and give less emphasis on data preprocessing 
and its impact on the final detection model. The authors stressed 
the importance of data preprocessing for Deepfake detections. 
Joshual \etal~\cite{Authors09} focused on the generality of 
facial forgery detection and found that most proposed systems 
lacked generality. The authors defined generality as reliably 
detecting multiple spoofing techniques and reliably spoofing 
unseen detection techniques.
 
Umur \etal~\cite{Authors08} proposed a generalized 
Deepfake detector called FakeCatcher using biological signals 
(internal representations of image generators and synthesizers). 
They used a simple Convolutional Neural Network (CNN) classifier 
with only three layers. The authors used 3000 videos for training 
and testing. However, they didn’t specify in detail how they 
preprocessed their data. From
~\cite{Authors15,Authors16,Authors17}, it is evident 
that very deep CNNs have superior performance than shallow 
CNNs in image classification tasks. Hence, there is still room
for another generalized Deepfake detector that has extensive 
data preprocessing pipeline and also is trained on a very deep 
Neural Network model to catch as many Deepfake artifacts 
as possible.

Therefore, we propose a generalized Convolutional Vision 
Transformer (CViT) architecture to detect Deepfake videos 
using Convolutional Neural Networks and the Transformer 
architecture. We call our approach generalized for three main 
reasons. 1) Our proposed model can learn local and global 
image features using the CNN and the Transformer architecture 
by using the attention mechanism of the Transformer~\cite{Authors63}. 2) We 
give equal emphasis on our data preprocessing during 
training and classification. 3) We propose to train our 
model on a diverse set of face images using the largest 
dataset currently available to detect Deepfakes created in 
different settings, environments, and orientations.

\section{Related  Work}

With the rapid advancement of the 
CNNs~\cite{Authors18,Authors19}, Generative 
Adversarial Networks (GANs)~\cite{Authors20}, 
and its variants~\cite{Authors21}, it is now 
possible to create hyper-realistic images~\cite{Authors22}, 
videos~\cite{Authors23} and audio signals
~\cite{Authors24,Authors25} that are much 
harder to detect and distinguish from real untampered 
audiovisuals. The ability to create a seemingly real 
sound, images, and videos have caused a steer from various 
concerned stakeholders to deter such developments not to be 
used by adversaries for malicious purposes~\cite{Authors01}. 
To this effect, there is currently an urge in the research 
community to come with Deepfake detection mechanisms. 

\subsection{Deep Learning Techniques for Deepfake 
Video Generation}

Deepfake is generated and synthesized by deep generative 
models such GANs and Autoencoders (AEs)
~\cite{Authors20,Authors26}. Deepfake is created 
by swapping between two identities of subjects in an image 
or video~\cite{Authors27}. Deepfake can also be 
created by using different techniques such as face 
swap~\cite{Authors28}, puppet-master~\cite{Authors24}, 
lip-sync~\cite{Authors29,Authors30}, 
face-reenactment~\cite{Authors31}, synthetic image 
or video generation, and speech synthesis~\cite{Authors32}. 
Supervised~\cite{Authors33,Authors34,Authors35}, 
and unsupervised image-to-image 
translation~\cite{Authors36} and video-to-video 
translation~\cite{Authors37,Authors38} can be 
used to create highly realistic Deepfakes.

The first Deepfake technique is the FakeAPP~\cite{Authors39} 
which used two AE network. An AE is a Feedforward 
Neural Network (FFNN) with an encoder-decoder 
architecture that is trained to reconstruct its input 
data~\cite{Authors40}. FakeApp’s encoder extracts 
the latent face features, and its decoder reconstructs the 
face images. The two AE networks share the same encoder 
to swap between the source and target faces, and 
different decoders for training.

Most of the Deepfake creation mechanisms focus on 
the face region in which face swapping and pixel-wise 
editing are commonly used~\cite{Authors03}. 
In the face swap, the face of a source image is swapped 
on the face of a target image. In puppet-master, 
the person creating the video controls the person in 
the video. In lip-sync, the source person controls the 
mouse movement in the target video, and in face 
reenactment, facial features are manipulated
~\cite{Authors27}. The Deepfake creation 
mechanisms commonly use feature map 
representations of a source image and target image. 
Some of the feature map representations are the 
Facial Action Coding System (FACS), image segmentation, 
facial landmarks, and facial boundaries~\cite{Authors26}. 
FACS is a taxonomy of human facial expression 
that defines 32 atomic facial muscle actions named 
Action Units (AU) and 14 Action Descriptors (AD) 
for miscellaneous actions. Facial land marks are a 
set of defined positions on the face, such as eye, 
nose, and mouth positions~\cite{Authors41}.

\subsubsection{Face Synthesis}

Image synthesis deals with generating unseen images 
from sample training examples~\cite{Authors42}. 
Face image synthesis techniques are used in face aging, 
face frontalization, and pose guided generation. 
GANs are used mainly in face synthesis. GANs are 
generative models that are designed to create 
generative models of data from samples
~\cite{Authors43,Authors20}. GANs contain 
two adversarial networks, a generative model $\mathit{G}$ , 
and discriminative model $\mathit{D}$ . The generator and 
the discriminator act as adversaries with respect 
to each other to produce real-like samples
~\cite{Authors21}. The generator’s goal is 
to capture the data distribution. The goal of the 
discriminator is to determine whether a sample 
is from the model distribution or the data distribution
~\cite{Authors20}. Face frontalization GANs 
change the face orientation in an image. Pose 
guided face image generation maps the pose of an 
input image to another image. GAN architecture, 
such as StyleGAN~\cite{Authors44} and 
FSGAN~\cite{Authors28}, synthesize highly 
realistic-looking images.

\subsubsection{Face Swap}

Face swap or identity swap is a GAN based method that 
creates realistic Deepfake videos. The face swap 
process inserts the face of a source image in a 
target image of which the subject has never 
appeared~\cite{Authors27}. It is most 
popularly used to insert famous actors in a 
variety of movie clips~\cite{Authors45}. 
Face swaps can be synthesized using GANs and 
traditional CV techniques such as FaceSwap 
(an application for swapping faces) and ZAO 
(a Chines mobile application that swaps anyone’s 
face onto any video clips)~\cite{Authors27}. 
Face Swapping GAN (FSGAN)~\cite{Authors28}, 
and Region-Separative GAN (RSGAN)~\cite{Authors46} 
are used for face swapping, face reenactment, 
attribute editing, and face part synthesis. Deepfake 
FaceSwap uses two AEs with a shared encoder that 
reconstructs training images of the source and 
target faces~\cite{Authors27}. The processes involve a face 
detector that crops and aligns the face using facial 
landmark information~\cite{Authors47}. 
A trained encoder and decoder of the source 
face swap the features of the source image 
to the target face. The autoencoder output is 
then blended with the rest of the image using 
Poisson editing~\cite{Authors47}. 

Facial expression (face reenactment) swap 
alters one’s facial expression or transforms 
facial expressions among persons. Expression 
reenactment turns an identity into a puppet
~\cite{Authors26}. Using facial expression 
swap, one can transfer the expression of a person 
to another one~\cite{Authors48}. Various facial 
reenactments have proposed through the years. 
CycleGAN is proposed by Jun-Yan \etal~\cite{Authors49} 
for facial reenactment between two video 
sources without any pair of training examples. 
Face2Face manipulates the facial expression 
of a source image and projects onto another 
target face in real-time~\cite{Authors50}. 
Face2Face creates a dense reconstruction 
between the source image and the target image 
that is used for the synthesis of the face images 
under different light settings~\cite{Authors47}.

\subsection{Deep Learning Techniques for Deepfake Video Detection}

Deepfake detection methods fall into three categories
~\cite{Authors51,Authors26}. Methods in the 
first category focus on the physical or psychological 
behavior of the videos, such as tracking eye blinking 
or head pose movement. The second category focus 
on GAN fingerprint and biological signals found in images, 
such as blood flow that can be detected in an image. 
The third category focus on visual artifacts. Methods 
that focus on visual artifacts are data-driven, and 
require a large amount of data for training. 
Our proposed model falls into the third category. 
In this section, we will discuss various architectures 
designed and developed to detect visual artifacts of Deepfakes.

Darius \etal~\cite{Authors06} proposed a CNN 
model called MesoNet network to automatically 
detect hyper-realistic forged videos created using 
Deepfake~\cite{Authors52} and Face2Face
~\cite{Authors50}. The authors used two 
network architectures (Meso-4 and MesoInception-4) 
that focus on the mesoscopic properties of an image. 
Yuezun and Siwei~\cite{Authors51} proposed
a CNN architecture that takes advantage of the image 
transform (i.e., scaling, rotation and shearing) 
inconsistencies created during the creation of 
Deepfakes. Their approach targets the artifacts 
in affine face warping as the distinctive feature to 
distinguish real and fake images. Their method 
compares the Deepfake face region with that of 
the neighboring pixels to spot resolution inconsistencies 
that occur during face warping.

Huy \etal~\cite{Authors53} proposed a novel 
deep learning approach to detect forged images 
and videos. The authors focused on replay attacks, 
face swapping, facial reenactments and fully computer 
generated image spoofing. Daniel Mas Montserrat 
\etal~\cite{Authors47} proposed a system 
that extracts visual and temporal features from 
faces present in a video. Their method combines a 
CNN and RNN architecture to detect Deepfake videos.

Md. Shohel Rana and Andrew H. Sung~\cite{Authors54}  
proposed a DeepfakeStack, an ensemble method 
(A stack of different DL models) for Deepfake detection. 
The ensemble is composed of XceptionNet, InceptionV3, 
InceptionResNetV2, MobileNet, ResNet101, 
DenseNet121, and DenseNet169 open source DL 
models. Junyaup Kim \etal~\cite{Authors55} 
proposed a classifier that distinguishes target individuals 
from a set of similar people using ShallowNet, VGG-16, 
and Xception pre-trained DL models. The main 
objective of their system is to evaluate the 
classification performance of the three DL models.

\section{Convolutional Vision Transformer}

\begin{figure*}[t]
\begin{center}
   \includegraphics[width=0.8\linewidth]{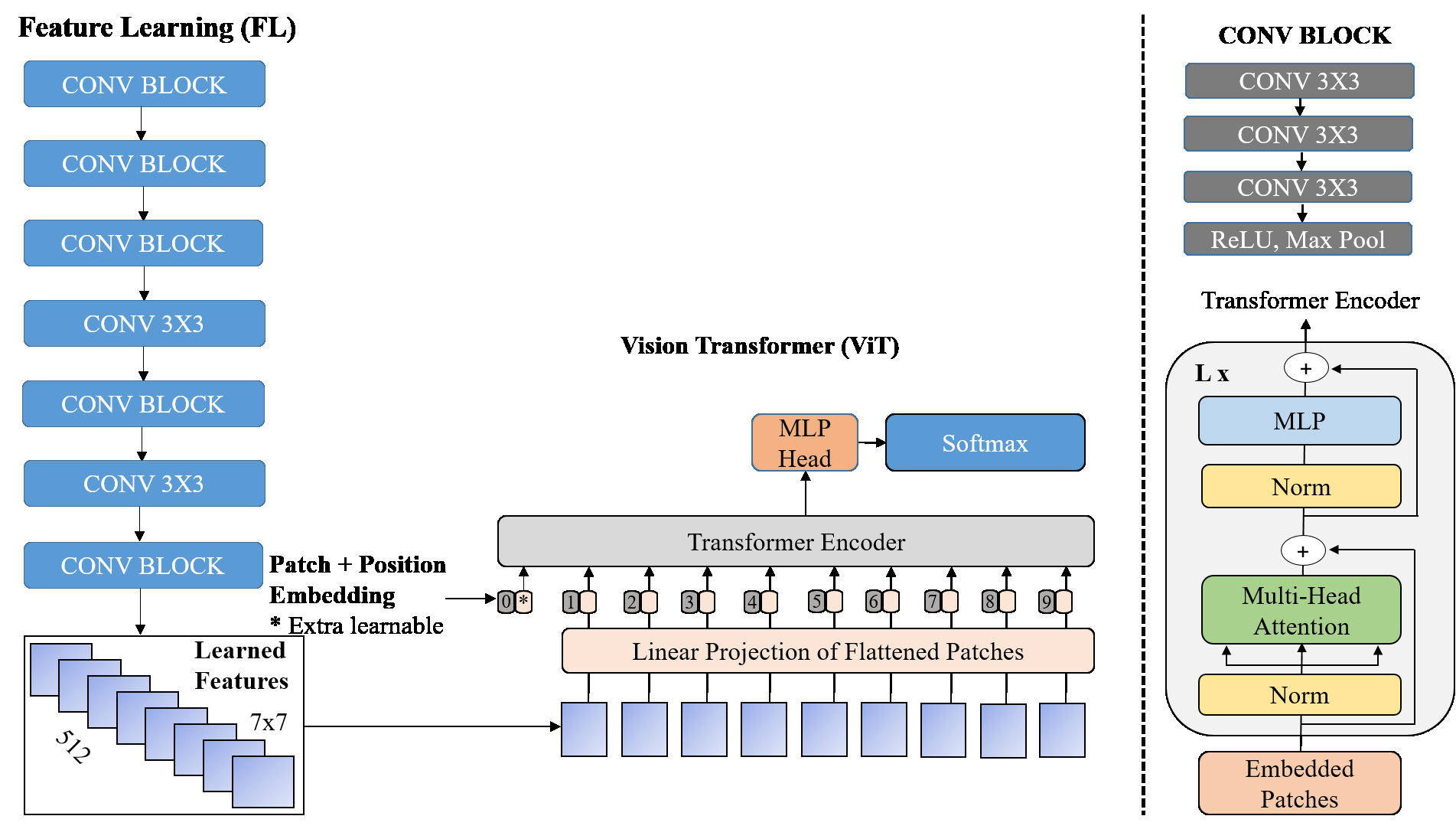}
\end{center}
   \caption{Convolutional Vision Transformer.}
\label{fig:cvit}
\end{figure*}

In this section, we present our approach to detect 
Deepfake videos. The Deepfake video detection model 
consists of two components: the preprocessing component 
and the detection component. The preprocessing 
component consists of the face extraction and data 
augmentation. The detection components consist of 
the training component, the validation component, 
and the testing component. The training and validation 
components contain a Convolutional Vision Transformer 
(CViT). The CViT has a feature learning component that 
learns the features of input images and a ViT 
architecture that determines whether a specific video 
is fake or real. The testing component applies the 
CViT learning model on input images to detect 
Deepfakes. Our proposed model is shown in Figure~\ref{fig:cvit}.

\subsection{Preprocessing}

\begin{figure}
\begin{center}
   \includegraphics[width=0.7\linewidth]{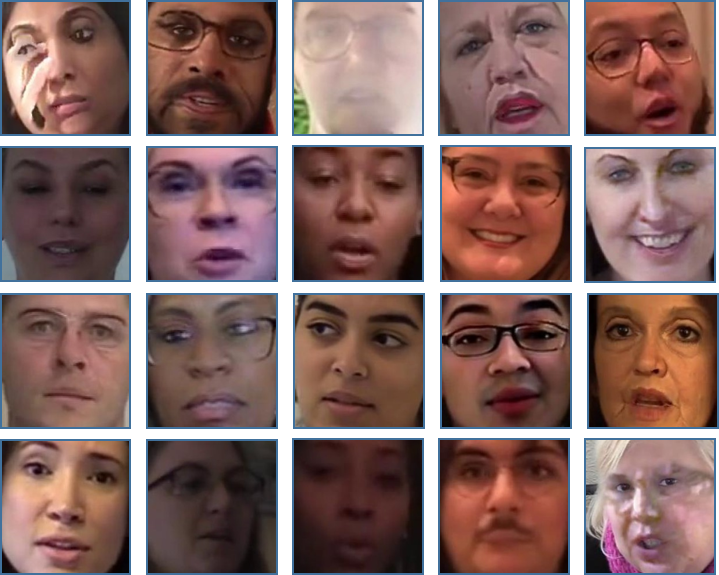}
\end{center}
   \caption{Sample extracted fake face images.}
\label{fig:fake}
\end{figure}

\begin{figure}
\begin{center}
   \includegraphics[width=0.7\linewidth]{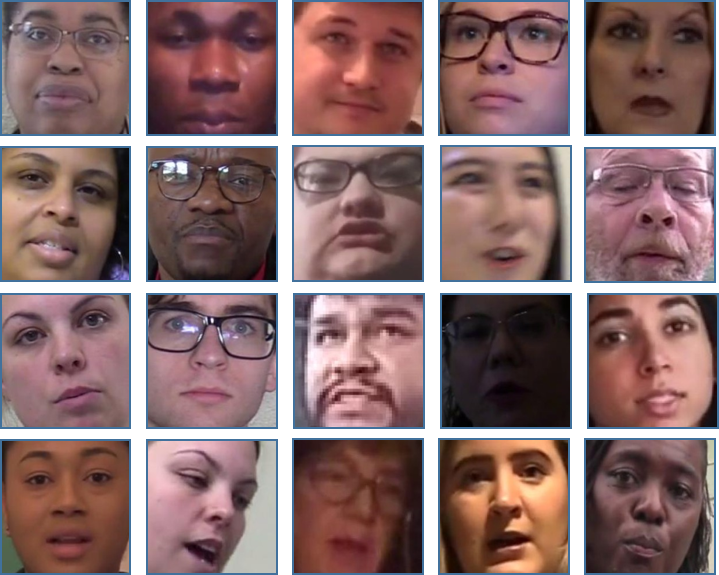}
\end{center}
   \caption{Sample extracted real face images.}
\label{fig:real}
\end{figure}

The preprocessing component’s function is to prepare 
the raw dataset for training, validating, and testing our 
CViT model. The preprocessing component has two 
sub-components: the face extraction, and the data 
augmentation component. The face extraction 
component is responsible for extracting face images 
from a video in a $\mathit{224}$ x $\mathit{224}$ RGB format. 
Figure~\ref{fig:fake} and Figure~\ref{fig:real} 
shows a sample of the extracted faces.

\subsection{Detection}

The Deepfake detection process consists of three 
sub-components: the training, the validation, and 
the testing components. The training component 
is the principal part of the proposed model. 
It is where the learning occurs. DL models require 
a significant time to design and fine-tune to fit a 
particular problem domain into its model. 
In our case, the foremost consideration is to search 
for an optimal CViT model that learns the features of 
Deepfake videos. For this, we need to search for the 
right parameters appropriate for training our dataset. 
The validation component is similar to that of the 
training component. The validation component is 
a process that fine-tunes our model. It is used to 
evaluate our CViT model and helps the CViT model 
to update its internal state. It helps us to track our 
CViT model’s training progress and its Deepfake 
detection accuracy. The testing component is 
where we classify and determine the class of 
the faces extracted in a specific video. Thus, 
this sub-component addresses our research 
objectives.

The proposed CViT model consists of two 
components: Feature Learning (FL) and the ViT. 
The FL extracts learnable features from the face 
images. The ViT takes in the FL as input and 
turns them into a sequence of image pixels for 
the final detection process.

The Feature Learning (FL) component is a 
stack of convolutional operations. The FL 
component follows the structure of VGG architecture
~\cite{Authors16}. The FL component differs 
from the VGG model in that it doesn’t have the fully 
connected layer as in the VGG architecture, and its 
purpose is not for classification but to extract face 
image features for the ViT component. Hence, the 
FL component is a CNN without the fully connected layer.

The FL component has 17 convolutional layers, with 
a kernel of $\mathit{3}$ x $\mathit{3}$. The convolutional layers extract 
the low level feature of the face images. 
All convolutional layers have a stride and 
padding of 1. Batch normalization to normalize
the output features and the ReLU activation function 
for non-linearity are applied in all of the layers. 
The Batch normalization function normalizes change
in the distribution of the previous layers
~\cite{Authors53}, as the change in between 
the layers will affect the learning process of the CNN 
architecture. A five max-pooling of a $\mathit{2}$ x $\mathit{2}$-pixel window 
with stride equal to 2 is also used. The max-pooling 
operation reduces dimension of image size by half. 
After each max-pooling operation, the width of the 
convolutional layer (channel) is doubled by a 
factor of 2, with the first layer having 32 
channels and the last layer 512.

The FL component has three consecutive 
convolutional operations at each layer, except 
for the last two layers, which have four 
convolutional operations. We call those three 
convolutional layers as CONV Block for simplicity. 
Each convolutional computation is followed by 
batch normalization and the ReLU nonlinearity. 
The FL component has 10.8 million learnable 
parameters. The FL takes in an image of size 
$\mathit{224}$ x $\mathit{224}$ x $\mathit{3}$, which is then convolved at each 
convolutional operation. The FL internal state 
can be represented as $\mathit{(C, H, W)}$ tensor, 
where $\mathit{C}$ is the channel, $\mathit{H}$ is the height, and 
$\mathit{W}$ is the width. The final output of the FL 
is a $\mathit{512}$ x $\mathit{7}$ x $\mathit{7}$ spatially correlated low 
level feature of the input images, which 
are then fed to the ViT architecture.

Our Vision Transformer (ViT) component 
is identical to the ViT architecture described 
in~\cite{Authors56}. Vision Transformer 
(ViT) is a transformer model based on the work 
of~\cite{Authors57}. The transformer 
and its variants (e.g., GPT-3~\cite{Authors58}) 
are predominantly used for NLP tasks. 
ViT extends the application of the transformer 
from the NLP problem domain to a CV problem 
domain. The ViT uses the same components as 
the original transformer model with slight 
modification of the input signal.
The FL component and the ViT component 
makes up our Convolutional Vision Transformer 
(CViT) model. We named our model CViT since 
the model is based on both a stack of 
convolutional operation and the ViT architecture.

The input to the ViT component is a feature 
map of the face images. The feature maps are 
split into seven patches and are then embedded 
into a $\mathit{1}$ x $\mathit{1024}$ linear sequence. The embedded 
patches are then added to the position embedding 
to retain the positional information of the image 
feature maps. The position embedding has a 
$\mathit{2}$ x $\mathit{1024}$ dimension.

The ViT component takes in the position 
embedding and the patch embedding and 
passes them to the Transformer. The ViT 
Transformer uses only an encoder, unlike 
the original Transformer. The ViT encoder 
consists of MSA and MLP blocks. The MLP 
block is an FFN. The Norm normalizes the 
internal layer of the transformer. The 
Transformer has 8 attention heads. The MLP 
head has two linear layers and the ReLU nonlinearity. 
The MLP head task is equivalent to the fully connected 
layer of a typical CNN architecture. The first layer 
has 2048 channels, and the last layer has two 
channels that represent the class of Fake or 
Real face image. The CViT model has a total of 
20 weighted layers and 38.6 million learnable 
parameters. Softmax is applied on the MLP 
head output to squash the weight values 
between 0 and 1 for the final detection purpose.

\section{Experiments}

In this section, we present the tools and 
experimental setup we used to design and 
develop the prototype to implement the model. 
We will present the results acquired from the 
implementation of the model and give an 
interpretation of the experimental results.

\subsection{Dataset }

DL models learn from data. As such, careful 
dataset preparation is crucial for their learning 
quality and prediction accuracy. BlazeFace neural 
face detector~\cite{Authors59}, MTCNN
~\cite{Authors60} and face\_recognition
~\cite{Authors61} DL libraries are used to 
extract the faces. Both BlazeFace and 
face\_recognition are fast at processing a 
large number of images. The three DL 
libraries are used together for added accuracy 
of face detection. The face images are stored 
in a JPEG file format with $\mathit{224}$ x $\mathit{224}$ 
image resolution. A 90 percent compression ratio 
is also applied. We prepared our datasets in 
a train, validation, and test sets. We used 162,174 
images classified into 112,378 for training, 24,898 
for validation and 24,898 for testing with $\mathit{70}$:$\mathit{15}$:$\mathit{15}$ 
ratios, respectively. Each real and fake class has 
the same number of images in all sets.

We used Albumentations for data augmentation. 
Albumentations is a python data augmentation 
library which has a large class of image transformations. 
Ninety percent of the face images were augmented, 
making our total dataset to be 308,130 facial images.

\subsection{Evaluation}

The CViT model is trained using the binary 
cross-entropy loss function. A mini-batch of 32 
images are normalized using mean of 
[$\mathit{0.485}$, $\mathit{0.456}$, $\mathit{0.406}$] and standard deviation 
of [$\mathit{0.229}$, $\mathit{0.224}$, $\mathit{0.225}$]. The normalized face 
images are then augmented before being fed into 
the CViT model at each training iterations. Adam 
optimizer with a learning rate of $\mathit{0.1e\text{-}3}$ and 
weight decay of $\mathit{0.1e\text{-}6}$ is used for optimization. 
The model is trained for a total of 50 epochs. 
The learning rate decreases by a factor of 0.1 
at each step size of 15.

The classification process takes in 30 facial 
images and passes it to our trained model. To 
determine the classification accuracy of our 
model, we used a log loss function. A log 
loss described in Equation \ref{eqn:logloss} classifies the 
network into a probability distribution from 
0 to 1, where $\mathit{0>y<0.5}$ represents the real 
class, and $\mathit{0.5\geq{y}<1}$ represents the fake class. 
We chose a log loss classification metric 
because it highly penalizes random guesses 
and confident false predictions.

\begin{equation}
LogLoss=\mathit{-\frac{1}n\sum_{i=1}^{n} [y_ilog(\hat{y}_i)+log(1-y_i)log(1-\hat{y}_i)]}
\label{eqn:logloss}
\end{equation}

Another metric we used to measure our model 
capacity is the ROC and AUC metrics~\cite{Authors62}. 
The ROC is used to visualize a classifier to select 
the classification threshold. AUC is an area covered 
by the ROC curve. AUC measures the accuracy of 
a classifier.

We present our result using accuracy, AUC score, 
and loss value. We tested the model on 400 
unseen DFDC videos and achieved 91.5 percent 
accuracy, an AUC value of 0.91, and a loss value 
of 0.32. The loss value indicates how far our 
models’ prediction is from the actual target 
value. For Deepfake detection, we used 30 face 
images from each video. The amount of frame 
number we use affects the chance of Deepfake 
detection. However, accuracy might not always 
be the right measure to detect Deepfakes as we 
might encounter all real facial images from a fake 
video (fake videos might contain real frames).

We compared our result with other Deepfake 
detection models, as shown in Table 
\ref{tab:ff}, \ref{tab:dfdc}, and \ref{tab:all}. 
From Table \ref{tab:ff}, \ref{tab:dfdc}, and \ref{tab:all}, 
we can see that our model performed well on the DFDC, 
UADFV, and FaceForensics++ dataset. However, 
our model performed poorly on the 
FaceForensics++ FaceShifter dataset. The 
reason for this is because visual artifacts are 
hard to learn, and our proposed model likely 
didn’t learn those artifacts well.

\begin{table}[htb]
\begin{center}
\begin{tabular}{|l|l|}
\hline
Dataset                             & Accuracy \\ \hline\hline
FaceForensics++   FaceSwap          & 69\%     \\ 
FaceForensics++   DeepFakeDetection & 91\%     \\
FaceForensics++   Deepfake          & 93\%     \\
FaceForensics++   FaceShifter       & 46\%     \\
FaceForensics++   NeuralTextures    & 60\%     \\ \hline
\end{tabular}
\end{center}
\caption{CViT model prediction accuracy on FaceForensics++ dataset}
\label{tab:ff}
\end{table}

\begin{table}[htb]
\begin{center}
\begin{tabular}{|l|ll|}
\hline
Method                   & Validation & Test    \\ \hline\hline
CNN and RNN-GRU~\cite{Authors47} {[}47{]} & 92.61\%    & 91.88\% \\ 
CViT                     & 87.25      & 91.5    \\ \hline
\end{tabular}
\end{center}
\caption{Accuracy of our model and other Deepfake detection models 
on the DFDC dataset}
\label{tab:dfdc}
\end{table}

\begin{table}[htb]
\begin{center}
\begin{tabular}{|l|lll|}
\hline
Method        & \multicolumn{1}{l|}{Validation} & \multicolumn{1}{l|}{FaceSwap} & \multicolumn{1}{l|}{Face2Face} \\ \hline\hline
MesoNet       & 84.3\%                          & 96\%                                          & \multicolumn{1}{l|}{92\%}                                           \\ 
MesoInception & 82.4\%                          & 98\%                                          & 93.33\%                                        \\ 
CViT          & 93.75                           & 69\%                                          & 69.39\%                                        \\ \hline
\end{tabular}
\end{center}
\caption{AUC performance of our model and other Deepfake detection models on UADFV dataset. * FaceForensics++}
\label{tab:all}
\end{table}

\subsection{Effects of Data Processing During Classification}

\begin{figure}[htb]
\begin{center}
   \includegraphics[width=4cm,height=5cm,keepaspectratio]{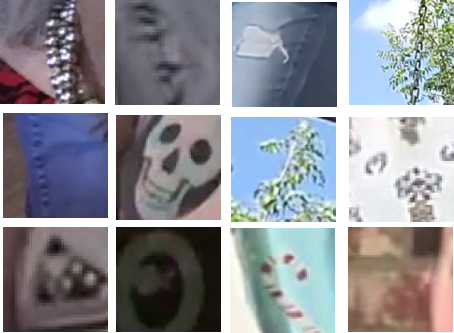}
\end{center}
   \caption{face\_recognition non face region detection.}
\label{fig:frec}
\end{figure}

\begin{figure}[htb]
\begin{center}
   \includegraphics[width=4cm,height=5cm,keepaspectratio]{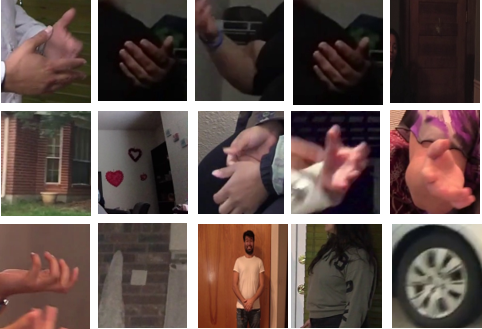}
\end{center}
   \caption{BlazeFace non face region detection.}
\label{fig:blaze}
\end{figure}

\begin{figure}[htb]
\begin{center}
   \includegraphics[width=4cm,height=5cm,keepaspectratio]{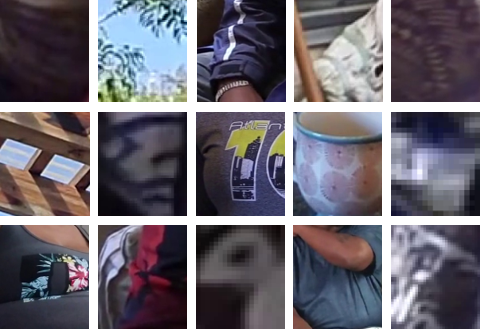}
\end{center}
   \caption{MTCNN non face region detection.}
\label{fig:mtcnn}
\end{figure}

A major potential problem that affects our model 
accuracy is the inherent problems that are in the 
face detection DL libraries (MTCNN, BlazeFace, 
and face\_recognition). Figure~\ref{fig:frec}, 
Figure~\ref{fig:blaze} and Figure~\ref{fig:mtcnn} 
show images that were misclassified by the 
DL libraries. The figures summarize our 
preliminary data preprocessing test on 200 
videos selected randomly from 10 folders. 
We chose our test set video in all settings 
we can found in the DFDC dataset: indoor, 
outdoor, dark room, bright room, subject sited, 
subject standing, speaking to side, speaking in 
front, a subject moving while speaking, gender, 
skin color, one person video, two people video, 
a subject close to the camera, and subject away 
from the camera. For the preliminary test, 
we extracted every frame of the videos and 
found the 637 nonface region.

We tested our model to check how its accuracy 
is affected without any attempt to remove these 
images, and our models’ accuracy dropped to 69.5 
percent, and the loss value increased to 0.4.

To minimize non face regions and prevent wrong 
predictions, we used the three DL libraries and 
picked the best performing library for our model, 
as shown in Table \ref{tab:libs}. As a solution, we used 
face\_recognition as a “filter” for the face 
images detected by BlazeFace. We chose 
face\_recognition because, in our investigation, 
it rejects more false-positive than the other two 
models. We used face\_recognition for final 
Deepfake detection.

\begin{table}[htb]
\begin{center}
\begin{tabular}{|l|lll|l|}
\hline
Dataset           & \multicolumn{1}{l|}{Blazeface} & \multicolumn{1}{l|}{f\_rec **} & \multicolumn{1}{l|}{MTCNN} \\ \hline\hline
DFDC              & 83.40\%                        & \textbf{91.50\%}                                & 90.25\%                    \\ 
FaceSwap          & 56\%                           & \textbf{69\%}                                   & 63\%                       \\ 
FaceShifter       & 40\%                           & \textbf{46\%}                                   & 44\%                       \\
NeuralTextures    & 57\%                           & \textbf{60\%}                                   & \textbf{60\%}                       \\
DeepFakeDetection & 82\%                           & \textbf{91\%}                                   & 79.59                      \\
Deepfake          & 87\%                           & \textbf{93\%}                                   & 81.63\%                    \\
Face2Face         & 54\%                           & 61\%                                   & \textbf{69.39\%}                    \\
UADF              & 74.50\%                        & \textbf{93.75\%}                                & 88.16\%                    \\\hline
\end{tabular}
\end{center}
\caption{DL libraries comparison on Deepfake detection accuracy. ** face\_recognition}
\label{tab:libs}
\end{table}

\section{Conclusion}

Deepfakes open new possibilities in digital media, 
VR, robotics, education, and many other fields. 
On another spectrum, they are technologies that 
can cause havoc and distrust to the general public. 
In light of this, we have designed and developed a 
generalized model for Deepfake video detection 
using CNNs and Transformer, which we named 
Convolutional Vison Transformer. We called our 
model a generalized model for three reasons. 1) 
Our first reason arises from the combined learning 
capacity of CNNs and Transformer. CNNs are strong 
at learning local features, while Transformers can 
learn from local and global feature maps. This 
combined capacity enables our model to correlate 
every pixel of an image and understand the 
relationship between nonlocal features. 2) 
We gave equal emphasis on our data preprocessing 
during training and classification. 3) We used the 
largest and most diverse dataset for Deepfake 
detection.

The CViT model was trained on a diverse collection 
of facial images that were extracted from 
the DFDC dataset. The model was tested 
on 400 DFDC videos and has achieved 
an accuracy of 91.5 percent. Still, our 
model has a lot of room for improvement. 
In the future, we intend to expand on our 
current work by adding other datasets 
released for Deepfake research to make 
it more diverse, accurate, and robust.

{\small
\bibliographystyle{ieee_fullname}
\bibliography{cvit}
}

\end{document}